\documentclass[conference]{IEEEtran}
\IEEEoverridecommandlockouts
\usepackage{cite}
\usepackage{amsmath,amssymb,amsfonts}
\usepackage{algorithmic}
\usepackage{graphicx}
\usepackage{textcomp}
\usepackage{xcolor}
\usepackage{multicol}
\usepackage{caption}
\usepackage{xfrac}
\usepackage[utf8]{inputenc}
\usepackage{caption,subcaption}
\def\BibTeX{{\rm B\kern-.05em{\sc i\kern-.025em b}\kern-.08em
    T\kern-.1667em\lower.7ex\hbox{E}\kern-.125emX}}
\begin{document}

\title{Automated System for Ship Detection from Medium Resolution Satellite Optical Imagery}

\author{\IEEEauthorblockN{Dejan Štepec}
\IEEEauthorblockA{XLAB Research \\
XLAB d.o.o. \\
Ljubljana, Slovenia \\
dejan.stepec@xlab.si}
\and
\IEEEauthorblockN{Tomaž Martinčič}
\IEEEauthorblockA{XLAB Research \\
XLAB d.o.o. \\
Ljubljana, Slovenia \\
tomaz.martincic@xlab.si}
\and
\IEEEauthorblockN{Danijel Skočaj}
\IEEEauthorblockA{Faculty of Computer and Information Science \\
University of Ljubljana \\
Ljubljana, Slovenia \\
danijel.skocaj@fri.uni-lj.si}
}

\maketitle

\begin{abstract}
In this paper we present a ship detection pipeline for low-cost medium resolution satellite optical  imagery obtained  from  ESA  Sentinel-2 and  Planet  Labs  Dove constellations. This optical satellite imagery  is  readily  available  for  any  place on Earth  and  underutilized in maritime domain, compared to existing solutions based on synthetic-aperture radar (SAR) imagery. We developed ship detection method based on state-of-the-art deep-learning based object detection method which was developed and evaluated on a large scale dataset that was collected and automatically annotated with the help of Automatic Identification System (AIS) data.
\end{abstract}


\section{Introduction}

Earth observations from space present a new dimension of information that offers an unprecedented global view for various domains and industries. Maritime industry is at the forefront of the utilisation of remote sensing data for ice, ship and oil pollution monitoring, with operational services available such as CleanSeaNet~\cite{CleanSeaNet} in Europe and NASTOP~\cite{NASTOP} in USA and Canada. Ship detection presents a crucial step in any maritime surveillance application in order to analyze traffic activity and also presents the main topic of this paper.

Synthetic-aperture radar (SAR) satellites represent the predominant source of remote sensing imagery for maritime surveillance, due to its resistance to weather conditions (e.g. clouds) and day \& night operational capabilities. SAR imagery has its limitations, especially near the coastline and in harbour areas due to presence of multiple objects on shore as well as on the sea, causing strong SAR backscatter centers and consequently the failure of traditional SAR ship detection methods~\cite{sar1}. Optical imagery presents a viable alternative for ship detection, increasing not only detection capabilities in complex environments, but also providing additional contextual information about the ship and its surroundings. 

Ship detection from optical imagery has become an active research area in recent years, especially with the advent of deep-learning based methods, which have greatly increased the performance and also led to first operational systems~\cite{optical_dlr_system}. Majority of research, as well as developed operational systems utilizing ship detection based on optical imagery, are focused on very-high-resolution (VHR) imagery, with spatial resolution up to 30cm. VHR satellites are available only in tasking modes which makes them expensive to collect imagery for specific areas of interest and longer periods of time, consequently limiting the use to governmental institutions for monitoring targeted limited areas in a specific time.

In this paper we present a ship detection pipeline for low-cost medium resolution satellite optical imagery, obtained from ESA Sentinel-2 (10m resolution) and Planet Labs Dove (3m resolution) constellations, that is readily available for any place on Earth and underutilized in the maritime domain. This satellite imagery is available free-of-charge in case of Sentinel-2 imagery or presents the cheapest commercial solution available on the market, with unique daily availability of new imagery for every point on Earth in the case of Planet Dove. To the best of our knowledge, this represents the first application of state-of-the-art deep-learning based methods for ship detection on this kind of satellite imagery in the research literature.

We adapt the state-of-the-art two-stage Mask R-CNN framework~\cite{mask_rcnn} to the domain of ship detection. Large scale datasets are needed to train the methods and there is no datasets available for medium resolution satellite imagery, as well as for our specific satellite constellations used. We utilize recently presented Kaggle Airbus ship detection dataset~\cite{airbus_challenge}. We show that these annotated datasets for ship detection on VHR satellite imagery can also be used for ship detection on a medium resolution imagery. To evaluate our proposed methods, we collect satellite imagery for the areas around Port of Oakland (San Francisco Bay) and Port of Long Beach for the years 2016 and 2017. In order to automatically annotate the data, we developed a procedure that combines openly available Automatic Identification System (AIS) data with obtained satellite imagery and cloud masks. With this approach we get the exact positional matching of AIS GPS data and satellite imagery, needed to evaluate ship detection performance. We also use AIS data in a novel way, such that we utilize AIS data to prepare weakly annotated ship detection dataset out of positional information and information about the ship length. With such a novel combination of existing VHR datasets and weakly annotated additional data, which can be obtained easily in large quantities, we report state-of-the-art detection results, as well as detection performance across different lengths. Our main contributions can be summarized as follows:

\begin{itemize}  
    \item We present a novel deep-learning-based state-of-the-art approach for ship detection from low-cost medium resolution satellite optical imagery with applications and evaluation on the off-the-shelf operational satellite imagery from ESA Sentinel-2 and Planet Dove constellations.
    
    \item We show that existing VHR annotated satellite imagery for ship detection can also be utilized on medium resolution imagery. Combined together with a generalized and automatic approach of building weakly annotated dataset from AIS data, makes this method applicable across different satellites with different sensor characteristics.
    
    \item We collect large scale satellite imagery from ESA Sentinel-2 and Planet Dove constellations and automatically combine it with AIS data to perform training and evaluation of the proposed methods on operational data as well as to enable additional future applications.
\end{itemize}

\section{Related Work}

SAR imagery is predominately used in maritime domain for ship detection. The most popular approaches are based on constant false alarm rate (CFAR) methods~\cite{cfar1, sumo}. With CFAR based methods, all pixels brighter than local threshold are regarded as pixels belonging to the ship. The threshold is computed, based on the local statistics and an assumed probability density function (PDF) for the clutter, as a pixel value above which a clutter pixel has a fixed probability of occurring~\cite{sumo}. Majority of the research work is focused on ship detection in open waters, omitting the need for reliable ship detection in port and harbour areas. In such areas, there is a presence of multiple objects on shore as well as on the sea, which are causing strong  SAR backscatter centers, which can cause a lot of false alarms with CFAR based methods~\cite{sar1}. Recently, there has been a spike in research towards making SAR based methods more reliable in such environments\cite{sar1,sar2,sar3}, majority of them still based on CFAR methods, but with a harbour specific sea-clutter models. We focus instead on optical satellite imagery, especially medium resolution, which is particularly underutilized in maritime domain~\cite{copernicus_catalogue} and can provide additional contextual information about the ship and its surroundings.

Ship detection can be in general terms of computer vision terminology viewed as an object detection problem. Computer vision domain has recently seen a tremendous improvements in performance in all  sorts  of  different  computer  vision  tasks, including object detection. This is  mostly  due  to  prevalence  of  deep-learning  based methods, that have  replaced  traditional  handcrafted  features  and learning  based classifiers with an end-to-end data-driven framework. This provides performance gains and improved  flexibility, by allowing the same methods to be successfully used in different tasks, given the availability of sufficient amount of labeled data. This is particularly the case of transfer learning, which is particularly important for the domains, where the data is scarce. Methods are pre-trained on large scale datasets for a common problem of image classification~\cite{imagenet} or object detection~\cite{coco} and then fine-tuned on smaller scale domain specific datasets.

Recently, there has been a spike in research towards using state-of-the-art deep-learning based object detection methods for ship detection on optical~\cite{optical_cnn_1, optical_cnn_2, hrsc2016}, as well as SAR imagery~\cite{sar_cnn_1, sar_cnn_2, sar_cnn_3}. Majority of them are using incarnations of Faster R-CNN~\cite{faster_rcnn} or YOLO~\cite{yolo} deep-learning based object detection methods. In this paper we also used an incarnation of Faster R-CNN framework with Feature Pyramid Network (FPN)~\cite{fpn} as a feature extractor. There is a lack of research for ship detection out of optical satellite imagery, especially when comparing with research in SAR domain. Majority of research for ship detection out of optical satellite imagery is also based on VHR optical imagery. We focus instead on lower resolution optical imagery from ESA Sentinel-2 and Planet Dove constellations.

The success of deep-learning based methods is extremely dependant on large amounts of labeled data, which is not always available, extremely difficult or expensive to obtain, or available with limited labeling accuracy. Ship detection datasets for optical satellite imagery are mainly captured from VHR imagery. In this paper we used Airbus Kaggle dataset~\cite{airbus_challenge}, which was provided in a 2018 organized Kaggle challenge, to detect ships out of the satellite imagery. This presents the biggest dataset of annotated ship locations (i.e. masks for rotated bounding boxes are provided). The dataset contains 150.000 JPEG images of 768x768 dimensions extracted from SPOT satellite which has 1.5m resolution. Most of the images do not contain ships, but negative examples. There are 81723 annotated ships; most of them are small ships, but there is also quite a bit of cargo ships present that can be filtered out. We have used scaling augmentations during the training process~\cite{high_low_res}, to perform domain adaptation for lower resolution satellite imagery, used in this paper. Besides using existing VHR ship detection datasets, we also utilized AIS data in a novel way, to automatically construct weakly annotated large scale ship detection dataset, with the help of positional and ship length information. Besides using AIS data for evaluation purposes~\cite{ais_bench_1}, this presents a novel way of using AIS data and eliminates the need of human level annotations and also presents a step in the direction of going beyond traditional supervised learning~\cite{weakly_1, weakly_2}.

\section{PIXSAT dataset}

\begin{figure*}[htbp]
    \centering
    \begin{subfigure}{0.47\textwidth}
      \includegraphics[width=\linewidth]{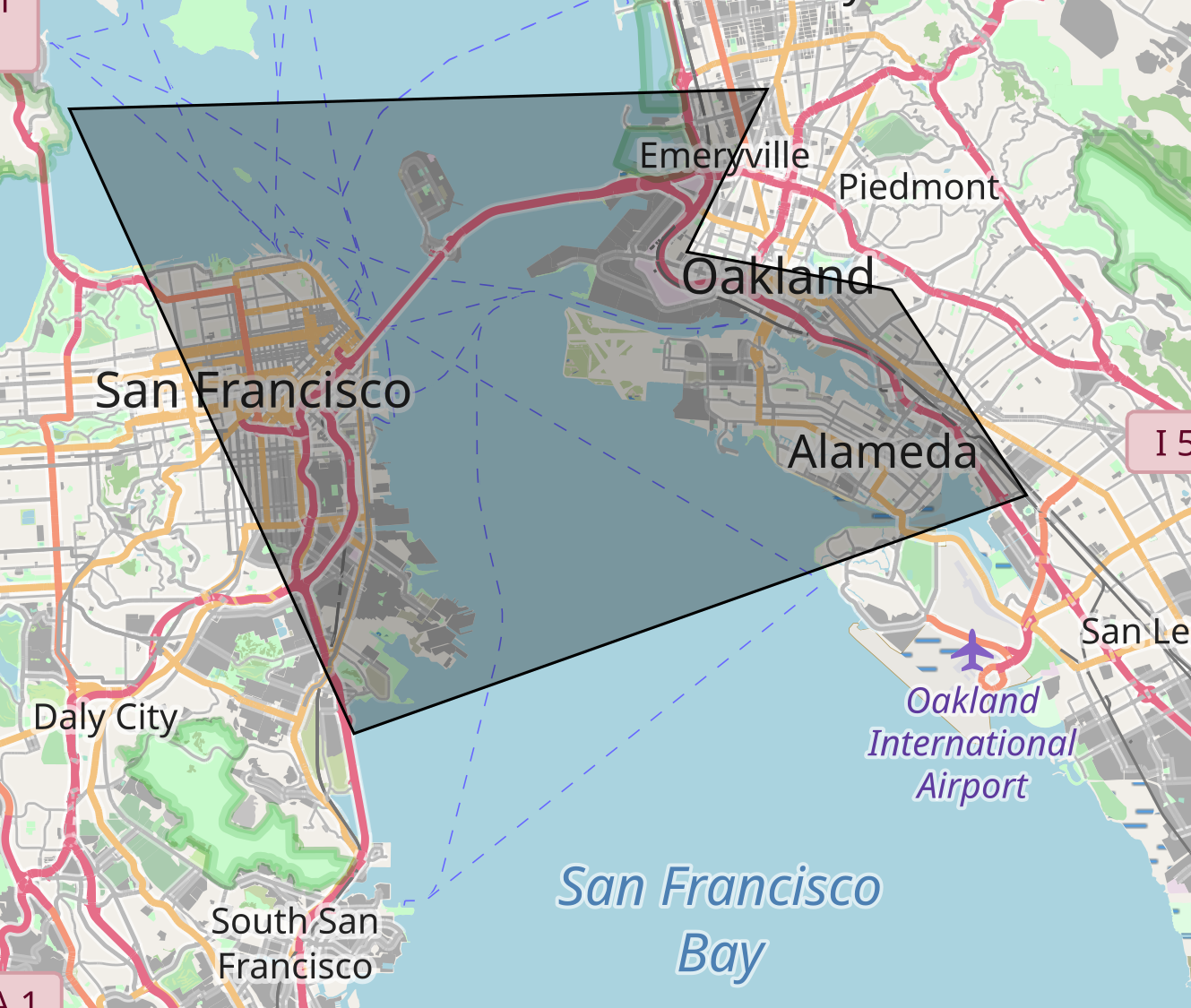}
      \caption{San Francisco Bay area}
      \label{fig:sub_sf_roi}
    \end{subfigure}\hfil
        \begin{subfigure}{0.47\textwidth}
      \includegraphics[width=\linewidth]{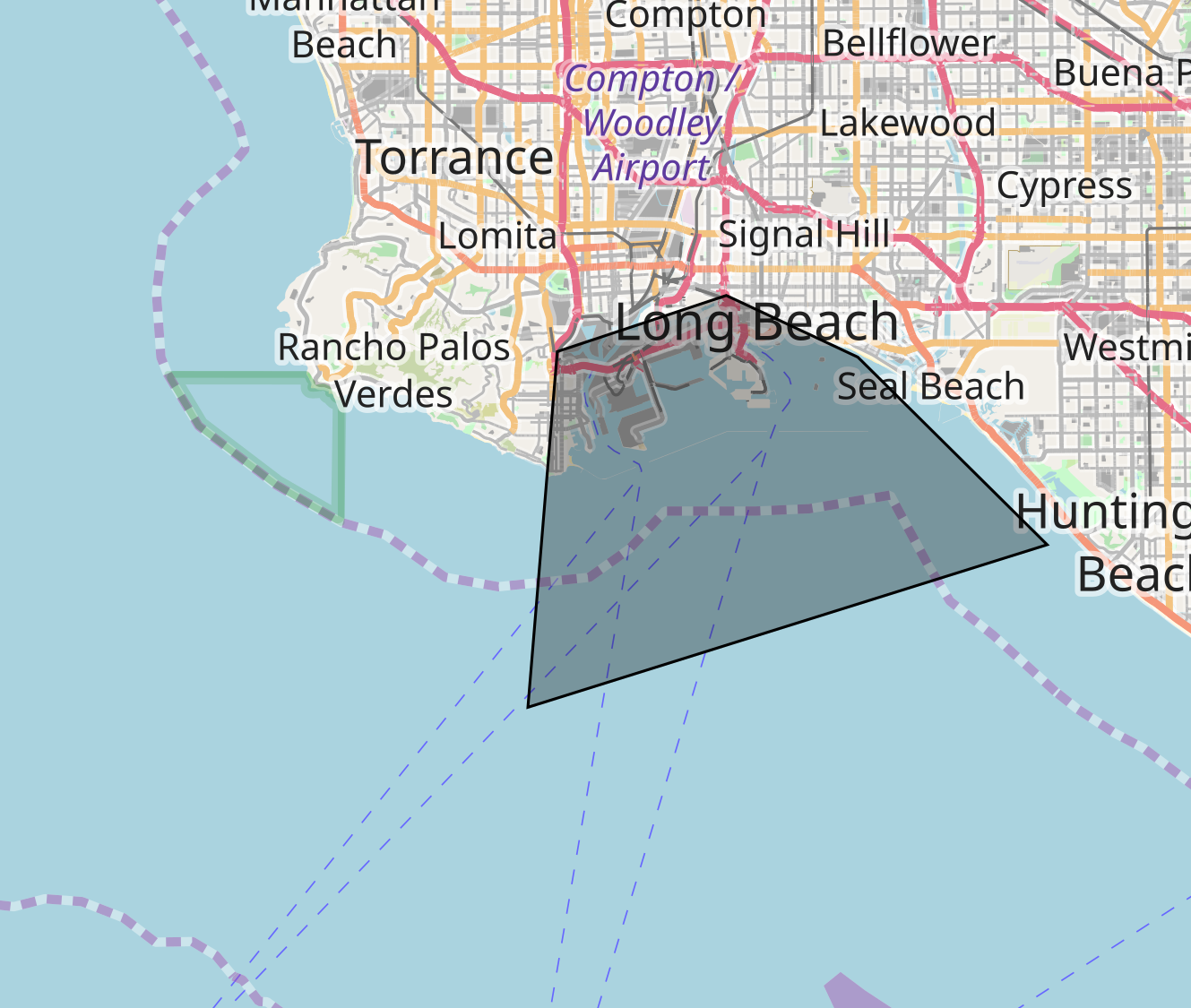}
      \caption{Port of Long Beach area}
      \label{fig:sub_lb_roi}
    \end{subfigure}\hfil
    \caption{Regions of interest (ROI) used for fetching satellite imagery.}
\end{figure*}

PIXSAT dataset\footnote{PIXSAT will be made available at: https://pixsat.xlab.si} a is large scale dataset for ESA Sentinel-2 and Planet Dove satellite constellations of optical imagery. Compared to existing datasets which are based mostly on VHR optical satellite imagery, this presents the first attempt to build a large scale dataset on medium resolution optical imagery. Existing large scale datasets such as HRSC2016~\cite{hrsc2016} are developed from Google Earth imagery, which does not represent real-life conditions due to selection of best possible imagery. Such satellite imagery, without any clouds and other image distortions due to sensors and different atmospheric conditions, is not realistically processed in an operational environment. Other large-scale datasets, such as Kaggle Airbus ship detection dataset~\cite{airbus_challenge}, are mostly captured on open sea, which greatly simplifies ship detection and does not represent significant gains over SAR imagery in terms of ship detection performance. PIXSAT dataset is compared to existing datasets, constructed from operational satellite imagery, capturing regions of Port of Oakland (San Francisco Bay) and Port of Long Beach for the years 2016 and 2017. We also developed a procedure to automatically combine AIS data from the ships with satellite imagery in order to automatically annotate ship positions in satellite imagery and to enrich them with metadata that is available in AIS messages.

In the next subsections we briefly describe different data sources utilized in the construction of the PIXSAT dataset and the methodology used to construct PIXSAT dataset.

\subsection{ESA Sentinel}

ESA Sentinels are a constellation of satellites with different sensors installed, which primary role is to ensure availability of Earth observation data for environmental and security services. The program is managed by the European Commission and implemented by European Space Agency (ESA), member states and other European agencies that rely on space data. The data from ESA is publicly available, including for commercial use. There
are five Sentinel missions, each mission is a constellation of two or more satellites in the same orbital plane, to have satisfying revisit time and coverage. The most important missions for maritime monitoring are Sentinel-1 and Sentinel-2 missions. Sentinel-1 mission (SAR) is already utilized by CleanSeaNet~\cite{CleanSeaNet}. We utilize Sentinel-2 imagery, which provides optical imagery, that is underutilized in maritime domain~\cite{copernicus_catalogue}.

Sentinel-2 mission offers a global coverage of all the land territory, as well as some of the maritime regions. All the coastlines are covered globally (20 km from the shore), as well as all inland water bodies, all closed seas and the whole Mediterranean Sea and is as such well suited to monitor coastal and harbour regions. Sentinel-2 mission revisit frequency at the equator is 5 days, with two satellites. Besides revisit times, spatial resolution is also important. Sentinel-2 imagery offers optical imagery with 10m resolution, which means that the minimum size of the object to be recognized is 10m.

Sentinel imagery can be accessed through ESA portals\footnote{https://scihub.copernicus.eu}, directly or via API access. More intuitive access is available through 3rd party providers such as Sinergise Sentinel Hub\footnote{https://www.sentinel-hub.com/}, that is also integrated into open-source Earth observation library eo-learn\footnote{https://eo-learn.readthedocs.io}. This library was used to retrieve imagery in this paper, along with additional capabilities such as cloud detection.

\subsection{Planet Dove}

Planet Labs is a commercial provider of satellite imagery and they emerged from smallsat revolution, as the biggest operator of optical satellites in space. They operate the so called cubesats, which are using commercial-off-the-shelf components in a much smaller form factor. Their cost is a fraction of the money required by traditional satellites and can thus be sent in the space in much larger constellations, offering much better revisit times. We have used Planet Dove constellation of 120+ satellites that are able to image entire Earth's landmass every day at 3m resolution.

Compared to ESA Sentinel-2, Planet Labs offers better spatial resolution and revisit times. They also offered (until recently) open access to California area, which was utilized in this paper. Increased capabilities offered by Planet constellation can complement the methods developed on ESA Sentinel imagery, as maritime area can be monitored with increased frequency and resolution which opens possibilities for additional use-cases. Planet Labs pricing is not publicly available, but studies have been done in agriculture, making it the most cost-effective solution among commercial providers~\cite{planet_cost}. Daily revisit times and increased resolution are unique features, not available by public (free-of-charge) providers.

Satellite imagery was retrieved from a freely available California area, licensed under CC-BY-SA 4.0, with available API. Planet also provides Usable Data Masks (UDM) which can be used for cloud masking. Recently new UDM-2\footnote{https://www.planet.com/pulse/planets-new-usable-data-masks/} was introduced which greatly improves the performance, but is only available on satellite imagery from August 2018 onwards. We were limited to historical satellite imagery from 2016 and 2017, due to availability of AIS data.

\subsection{Marine Cadastre AIS}

Automatic Identification System (AIS) was proposed and mandated by International Maritime Organization (IMO) and it's main intention is to prevent collisions on sea. It provides additional information, however, it does not replace
existing solutions on board, such as radar and other means that are regulated by Convention on the International Regulations for Preventing Collisions at Sea (COLREG). Since December 31st, 2004 all vessels exceeding 300GT are obligated to have an AIS transceiver installed and operational. Navigational data, information about the ship and voyage related data, is transmitted via VHF radio between ships and shore stations. The range is limited to the VHF range, which is about 10-20 nautical miles but S-AIS (Satellite-based AIS) is available, which can track ships on open sea. Besides collision avoidance, AIS data is used for many other applications in maritime domain, such as fishing fleet monitoring, maritime security, search and rescue, accident investigation, fleet and cargo tracking and many others. We utilize AIS data in a novel way, not only for evaluating ship detection performance, but also to build large-scale weakly annotated dataset for ship detection, eliminating the need for human level annotations.

Kinematic information (ship location, speed, course, heading, etc.) and some static
information (MMSI, ship type, ship size, etc.) are provided every couple of seconds when ship is underway and every couple of minutes when the ship is anchored or moored. This data is available in almost real-time, while historical data is also available. Marine Cadastre provides historical AIS data for USA coastal and inland waters\footnote{https://marinecadastre.gov/ais/}. We used the data from San Francisco Bay and Port of Long Beach areas. AIS data is updated once per year and currently the AIS data is available from 2009 to 2017.

\begin{figure*}[ht!]
\centering
\centerline{\includegraphics[width=1\textwidth]{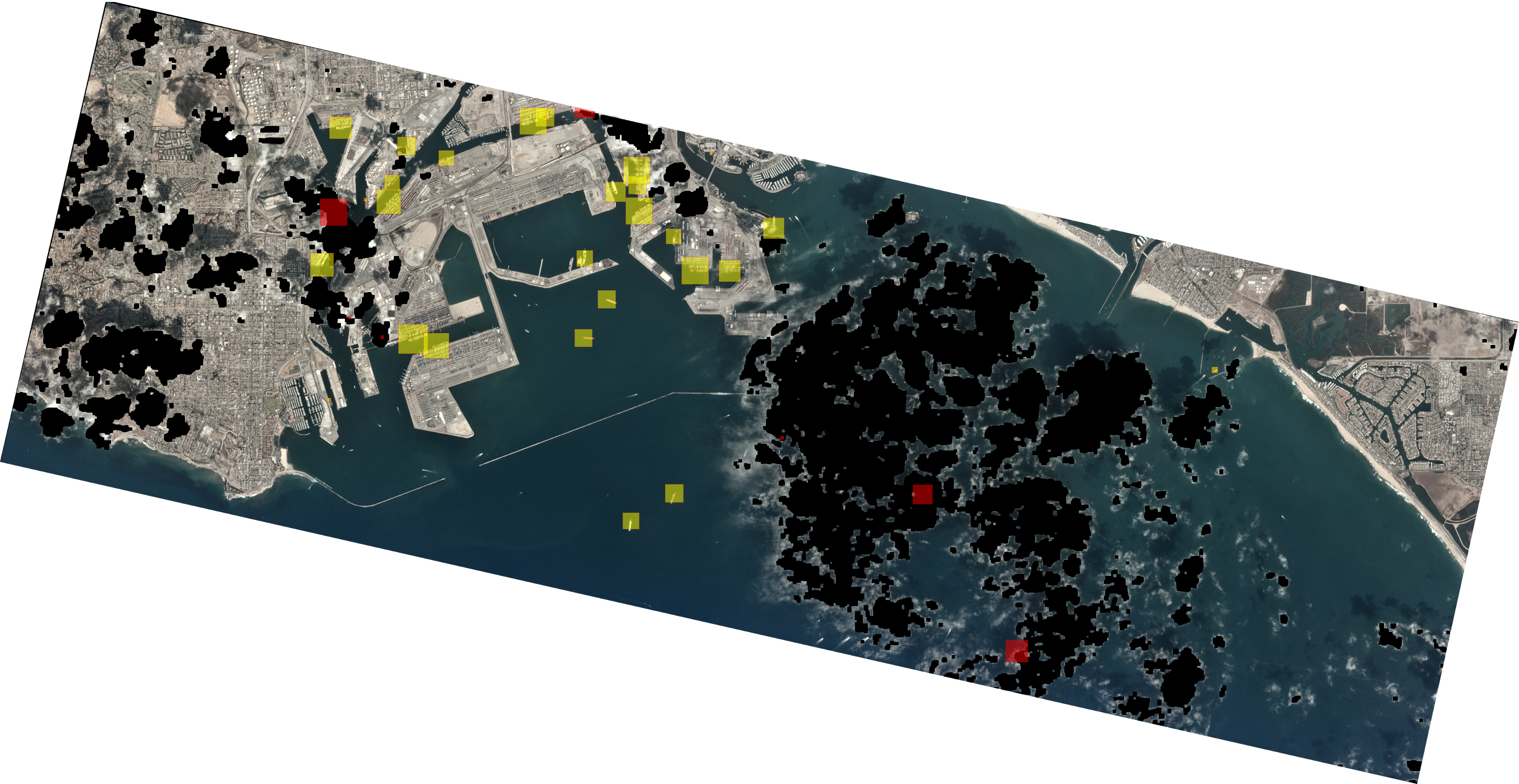}}
\caption{Satellite imagery of Port of Long Beach and matched AIS data. Cloud mask (UDM) is also visualized in black and rectangles covered in clouds by more than 20\% are visualized in red. Best viewed in digital version with zoom.}
\small\textsuperscript{*} Contains modified Open California Satellite Imagery \textcopyright 2019 Planet Labs Inc. licensed under CC BY-SA 4.0.
\label{fig:planet_lb_sample}
\end{figure*}

\subsection{Properties of PIXSAT}

PIXSAT dataset is composed out of ESA Sentinel-2 and Planet Dove satellite imagery. Satellite imagery was obtained for areas of Port of Long Beach and San Francisco Bay, for the years 2016 and 2017. Regions of interest (ROIs) that were used for fetching the imagery are presented in images~\ref{fig:sub_lb_roi} and~\ref{fig:sub_sf_roi}, for Port of Long Beach and San Francisco Bay respectively. Number of satellite images, that were obtained for each region and each year are presented in table~\ref{stat_images}. Altogether 2420 satellite images were obtained from Planet Dove and 148 from ESA Sentinel satellite constellations. The large difference in quantity is due to better revisit frequency of Planet Dove and due to Planet API ROI fetching strategy, which does not combine satellite imagery together but rather fetches all the raw imagery that intersect ROI area. There is also greater number of imagery for year the 2017, which is due to continuously larger Planet Dove constellations, which achieved daily landmass coverage at the end of the year 2017. Similarly, additional satellite was provided for Sentinel-2 constellation in the beginning of the 2017. Besides imagery, cloud masks (UDM) were also fetched from Planet API. We have used eo-learn library to obtain cloud masks for Sentinel-2 imagery. We have found out that the provided cloud masks from Planet are not of sufficient quality to rely on it. New version of UDM API, available for satellite imagery from August 2018 onwards, should provide better results.

\begin{table}[htbp]
\caption{PIXSAT statistics - number of images}
\begin{center}
\begin{tabular}{|c|c|c|c|c|}
\hline
 &\multicolumn{2}{|c|}{\textbf{SF}}&\multicolumn{2}{|c|}{\textbf{LB}} \\
\cline{2-5} 
 & \textbf{2016}& \textbf{2017}& \textbf{2016}& \textbf{2017} \\
\hline
\textbf{Planet} & 192/78 & 1000 & 212/117 & 1016 \\
\hline
\textbf{Sentinel} & 31/24 & 41 & 27/22 & 49 \\
\hline
\end{tabular}
\label{stat_images}
\end{center}
\end{table}

\begin{figure}[ht!]
\centerline{\includegraphics[width=0.5\textwidth]{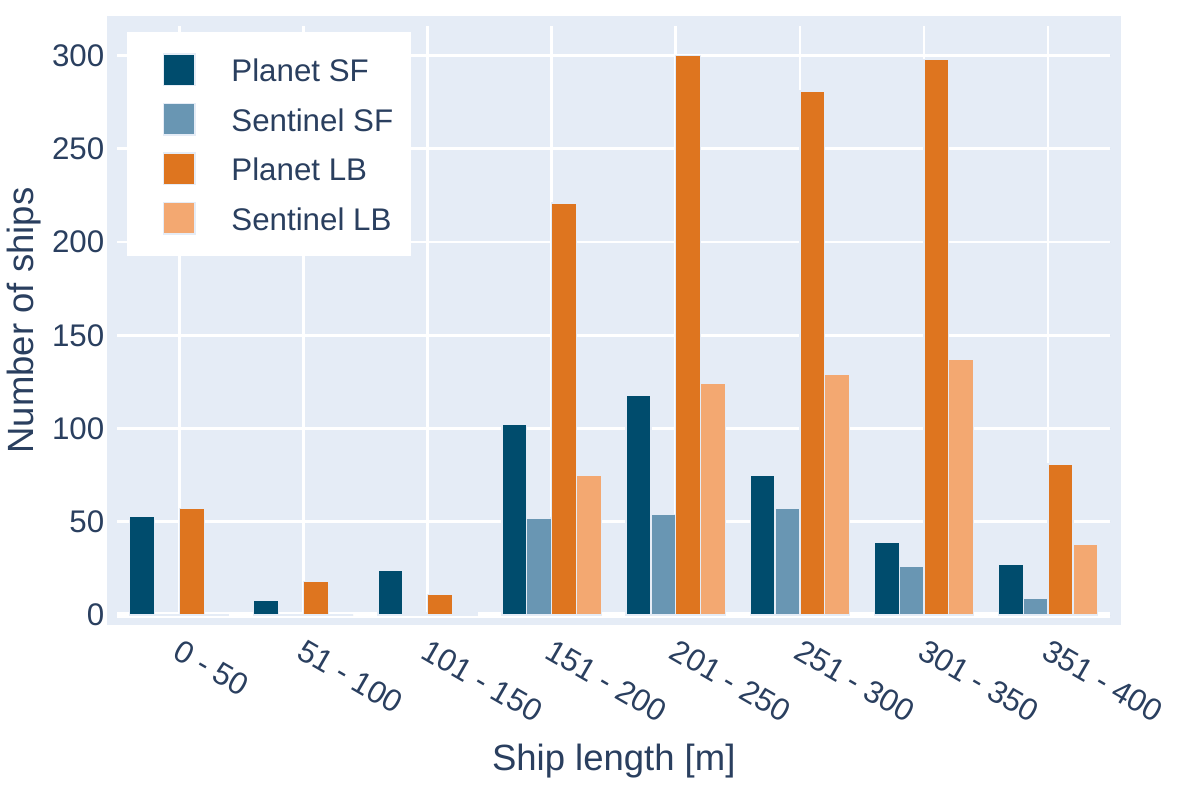}}
\caption{Histogram of reported AIS ship lengths for the ships captured in the 2016 part of the PIXSAT dataset, that was used for evaluation purposes.}
\label{fig:2016_length_hist}
\end{figure}

Satellite imagery was combined with AIS data in order to automatically annotate satellite imagery with ship positions and additional information, such us ship length and navigational status. The number of ships (as reported by AIS), that were correlated with satellite imagery for each region and year is presented in table~\ref{stat_ships}. Altogether 34894 ships were matched with Planet imagery and 5251 with Sentinel imagery. Only the ships of length greater than 30m were used for matching, due to our interest in commercial ships and spatial resolution limitations of used satellite imagery. We only matched ships whose reported navigational status in AIS messages is not "underway using engine", to have a direct positional matching, without the need of interpolation. We have used 5 minutes window for AIS positional report averaging of such stationary ships in order to ensure better positional accuracy. Matched ships might be covered with clouds or distorted due to different atmospheric conditions or image distortions.
The data from 2016 was used for evaluation and 2017 for training purposes. In order to have valid ground truth annotations, we manually inspected all the satellite imagery from 2016 and its AIS matches. The second number for the year 2016 in tables~\ref{stat_images} and~\ref{stat_ships} provides the number of images containing at least one valid AIS matching and the number of valid ships, respectively. We retained AIS matches that were not covered by clouds and can be clearly recognized by a human. An example of satellite imagery and AIS matches is presented in figure~\ref{fig:planet_lb_sample}, where rectangles are visualized around reported AIS positions with edge length corresponding to twice the reported AIS ship length. Rectangles that are covered in clouds for more than 20\% are visualized in red. Cloud detection from UDM masks did not prove reliable and because of that, resulting patches for the year 2016 were manually inspected. With better cloud detection methods, or with new, more reliable UDM masks, one can generate such dataset in a complete automatic way. In figure~\ref{fig:2016_length_hist} we present distribution of ship lengths as captured in the 2016 part of the dataset, that was used for evaluation.

\begin{table}[htbp]
\caption{PIXSAT statistics - number of ships}
\begin{center}
\begin{tabular}{|c|c|c|c|c|}
\hline
 &\multicolumn{2}{|c|}{\textbf{SF}}&\multicolumn{2}{|c|}{\textbf{LB}} \\
\cline{2-5} 
 & \textbf{2016}& \textbf{2017}& \textbf{2016}& \textbf{2017} \\
\hline
\textbf{Planet} & 2085/446 & 10481 & 4076/1267 & 18252 \\
\hline
\textbf{Sentinel} & 669/199 & 957 & 1229/504 & 2396 \\
\hline
\end{tabular}
\label{stat_ships}
\end{center}
\end{table}

An additional dataset was prepared out of the original PIXSAT dataset, that was used for training the ship detection method. Besides using AIS data for evaluating ship detection performance, we used ship length information to provide weakly annotated data for training ship detection method. We made patches out of the original satellite imagery of sizes 800x800 pixels, with 200 pixels of overlap and merged it with AIS data. Rectangles with the reported ship length were centered around the reported ship positions. We manually inspected all the 800x800 patches with annotations, to discard patches covered with clouds or patches where some of the AIS matches were false. The number of patches and the number of annotated ships for each data provider, location and year are reported in table~\ref{stat_ships_detection}. Annotations on a full Planet Dove satellite (before patching) are presented in figure~\ref{fig:planet_lb_sample}. Only the data from 2017 was used for training purposes, as 2016 data was used for evaluation, as reported in tables~\ref{stat_images} and~\ref{stat_ships}.

\begin{table}[htbp]
\caption{PIXSAT statistics - number of patches and ships}
\begin{center}
\begin{tabular}{|c|c|c|c|c|}
\hline
 &\multicolumn{2}{|c|}{\textbf{SF}}&\multicolumn{2}{|c|}{\textbf{LB}} \\
\cline{2-5} 
 & \textbf{2016}& \textbf{2017}& \textbf{2016}& \textbf{2017} \\
\hline
\textbf{Planet} & 137/231 & 740/1417 & 601/1282 & 1773/5350 \\
\hline
\textbf{Sentinel} & 33/188 & 39/278 & 40/708 & 102/1629 \\
\hline
\end{tabular}
\label{stat_ships_detection}
\end{center}
\end{table}

\section{Experiments}

In this section we present the experimental setup and methodology, as well as the results of our automated system for ship detection. We report detection performance across different ship lengths, different locations and satellite constellations.

\subsection{Training data}

For training we utilized the large scale Kaggle Airbus dataset~\cite{airbus_challenge}, which presents the largest ship detection dataset in the research community. The dataset was modified such that bounding boxes were fitted on provided rotated masks, to translate initial segmentation problem to that of object detection. We also filtered out only annotations where provided lengths of initial rotated masks exceeded the length of 50m. This is due to a large amount of smaller ships present in the dataset, which would present noise in lower resolution satellite imagery, used in this paper. After filtering, we were left out with 26496 images with altogether 42803 ship annotations. Additionally we added the same amount of images without any annotations. Inspired by~\cite{high_low_res}, we also experimented with scale augmentations. For each image with annotations, we added additional one, scale augmented in the range of 50\% to 70\% of its original size. We also used rotation augmentation in the range of -45\textdegree to 45\textdegree and Gaussian blur in the range of 0 to 0.5.

We also used a novel weakly annotated PIXSAT dataset, which was presented in this paper. We use PIXSAT in an independent way and in a fine-tuning procedure in a combination with Kaggle Airbus dataset. No augmentation procedure was used with PIXSAT dataset.

\subsection{Methods}

We used the state-of-the-art two-stage Mask R-CNN framework~\cite{mask_rcnn}. More precisely, we used modified Faster R-CNN~\cite{faster_rcnn} implementation with Feature Pyramid Network~\cite{fpn} and ResNet-50~\cite{resnet} as a backbone architecture. We used the code from Facebook~\cite{mask_rcnn_code}, which provides fast and modular implementation in PyTorch. We trained the method on Kaggle Airbus dataset for 100 epochs with initial learning rate of $0.02$, which was reduced by a factor of 10 at \sfrac{1}{3} and \sfrac{2}{3} of epochs. When training on Kaggle Airbus dataset, network was initialized from ImageNet~\cite{imagenet}. We also performed training on PIXSAT dataset, initialized from ImageNet, as well from a model trained on Kaggle Airbus. When fine-tuning PIXSAT from Kaggle Airbus, we reduced the initial learning rate by a factor of 10. Confidence of 20\% was used for detection threshold.

Satellite images are divided in 800x800 patches with 200 pixels of overlap, due to resource constraints. Same parameters were also used to build PIXSAT ship detection part of the dataset. Kaggle Airbus dataset also has a similar resolution of 768x768 pixels. We used Non-Maximum Suppression procedure (NMS), to eliminate multiple overlapping detections.

\subsection{Experimental results}

\begin{figure*}[ht!]
\begin{multicols}{2}
    \includegraphics[width=\linewidth]{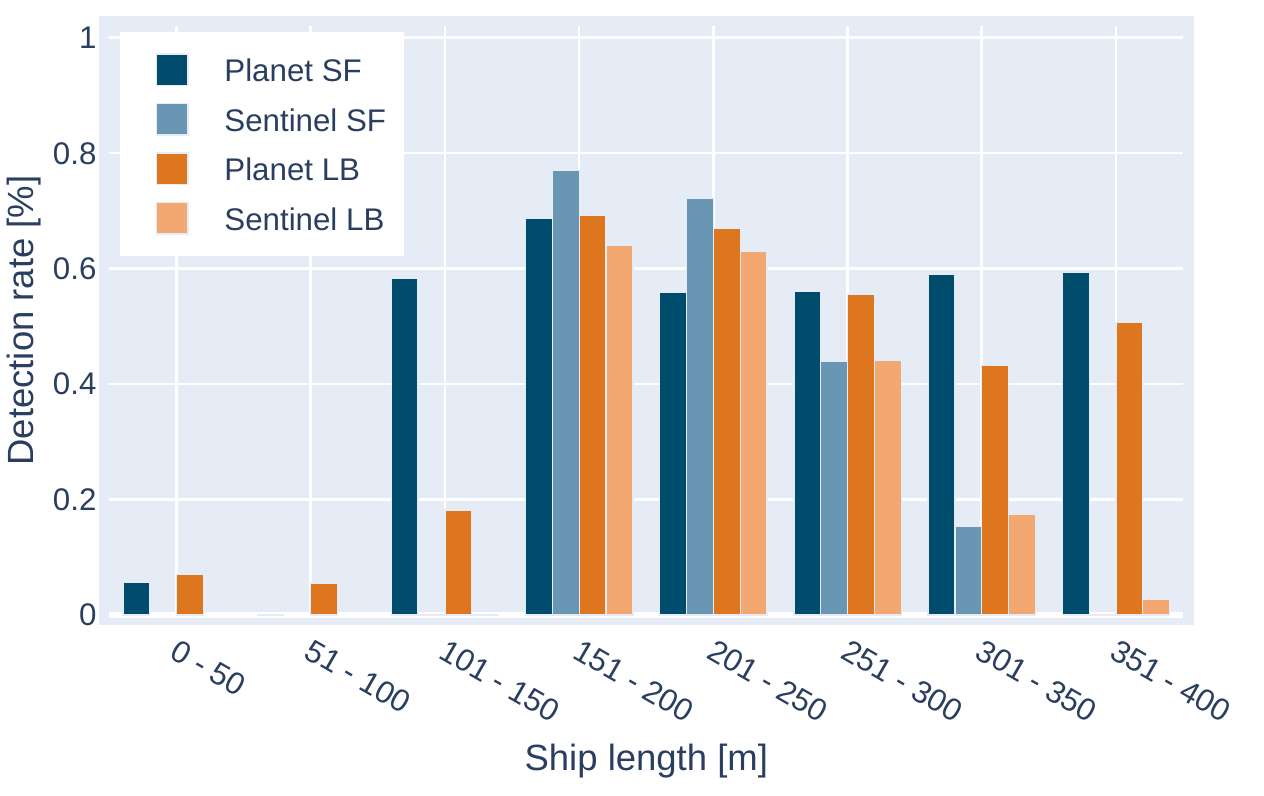}
    \captionof{figure}{Baseline - Kaggle Airbus}
    \label{baseline_hist}
    \par
    \includegraphics[width=\linewidth]{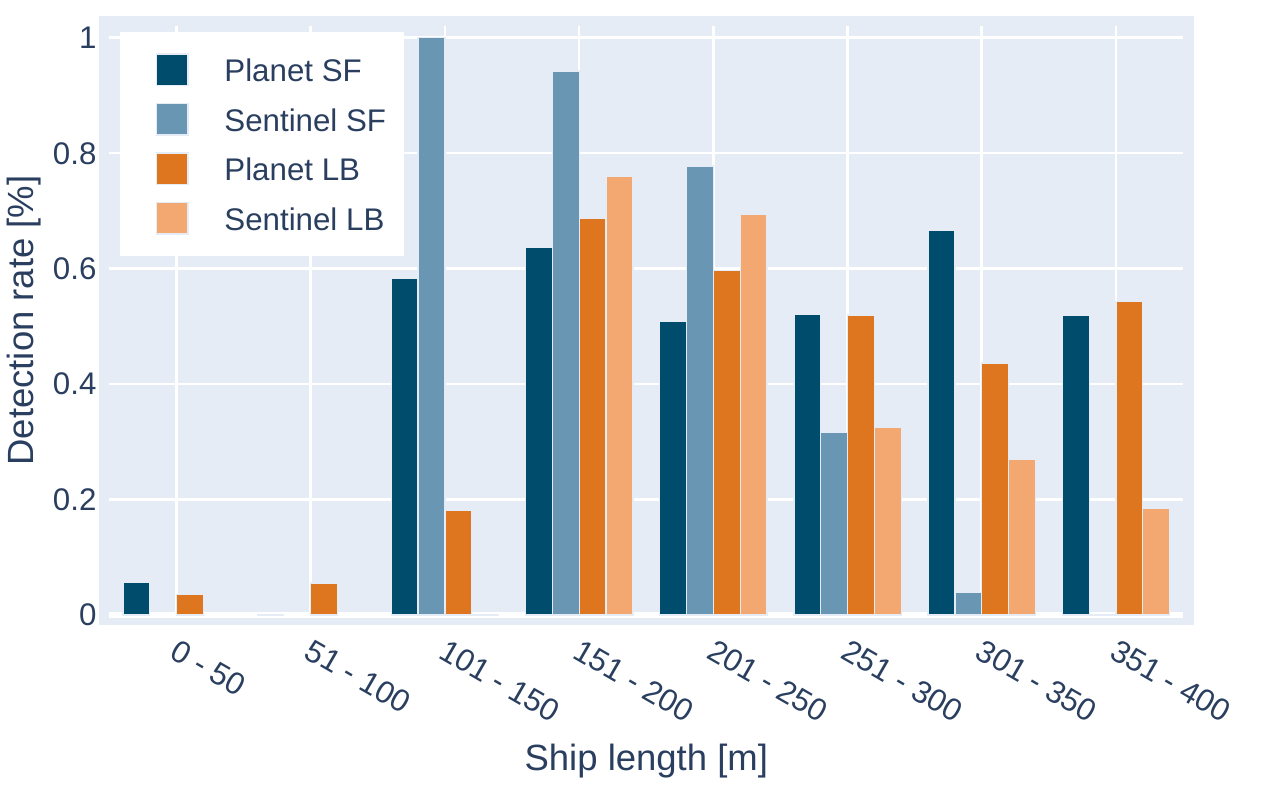}
    \captionof{figure}{Baseline - Kaggle Airbus (with augmentations)}
    \label{baseline_aug_hist}
    \par 
    \end{multicols}
\begin{multicols}{2}
    \includegraphics[width=\linewidth]{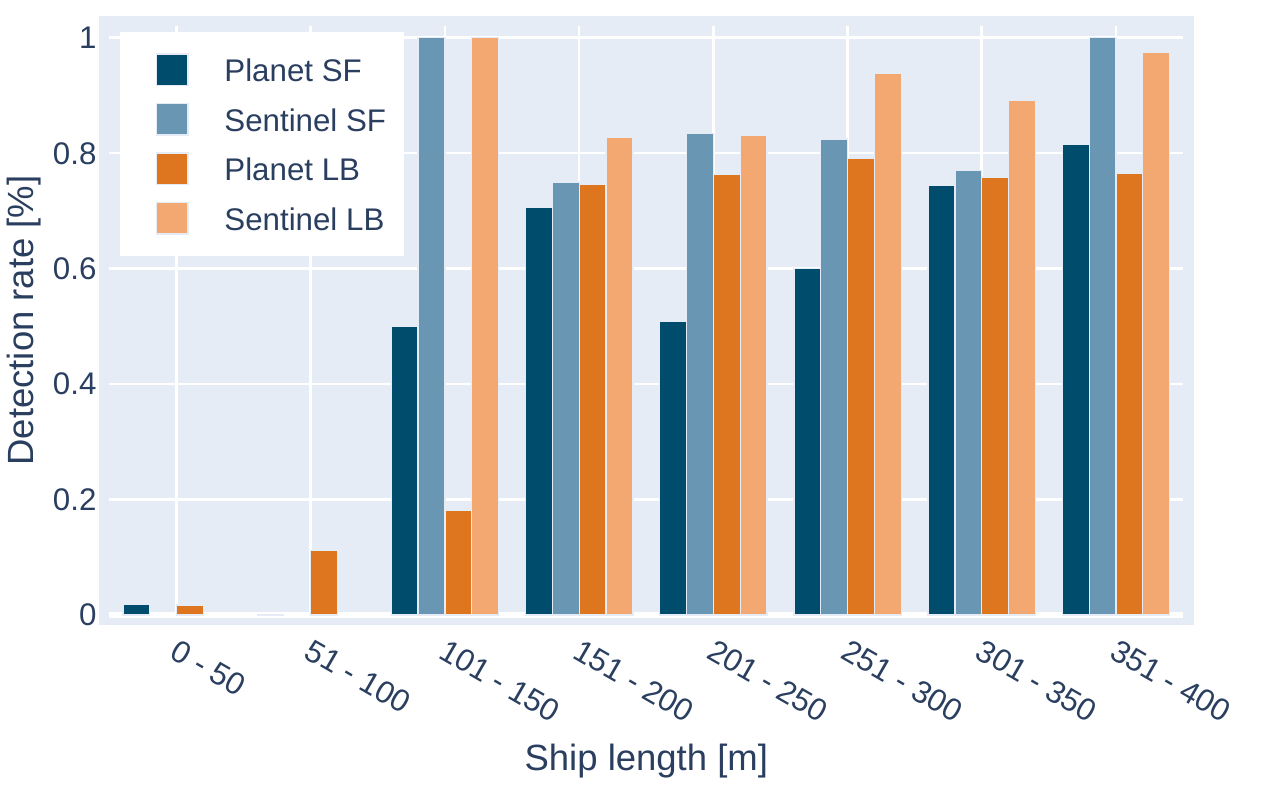}
    \captionof{figure}{PIXSAT only}
    \label{pixsat_hist}
    \par
    \includegraphics[width=\linewidth]{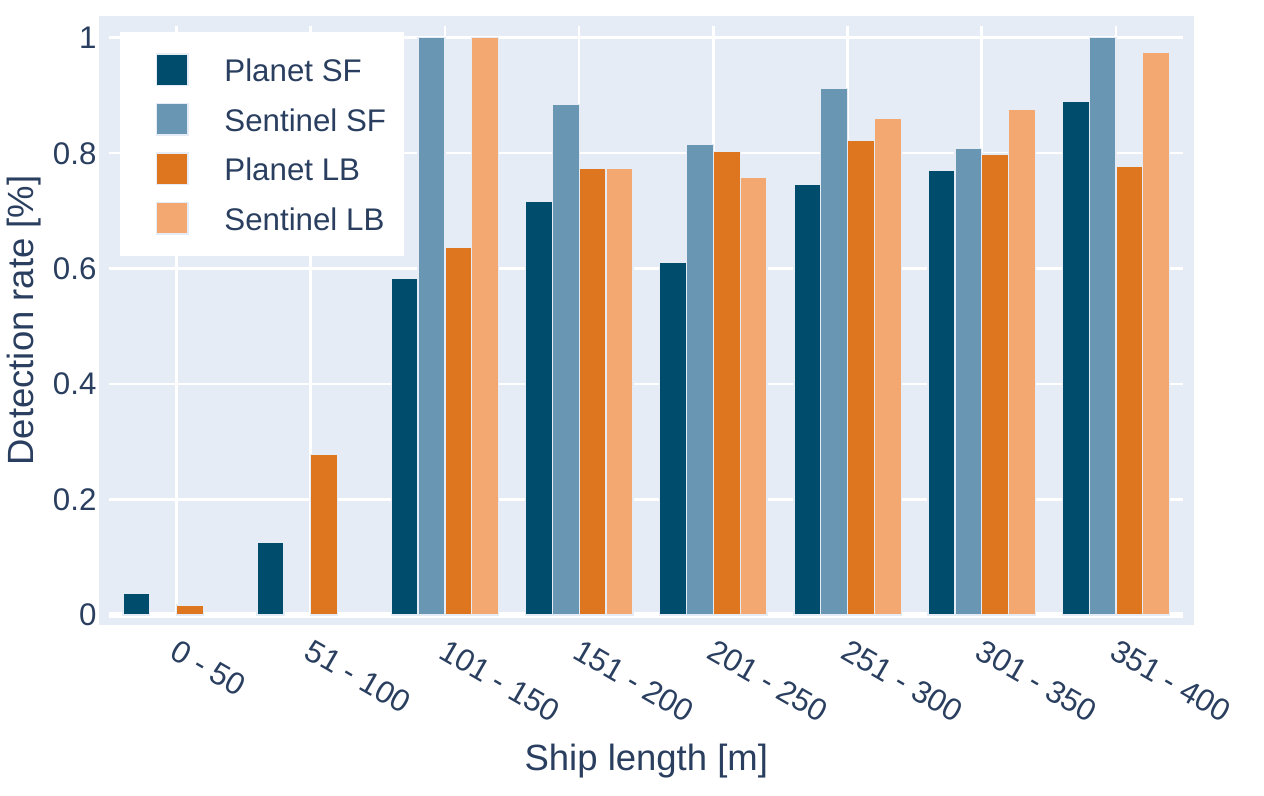}
    \captionof{figure}{PIXSAT fine-tuned on Kaggle Airbus (no aug.)}
    \label{pixsat_finetune_hist}
    \par
\end{multicols}
\end{figure*}

We evaluated our method on a 2016 part of the PIXSAT dataset. We report ship detection (retrieval) rate, which is considered successful if ground truth position from AIS data is inside the reported detection from our method. We report detection rates across different satellite constellations, different locations and different ship lengths. We conduct multiple experiments in order to test different augmentation and training strategies. 

In the first experiment (i.e. baseline), we train ship detection method only on Kaggle Airbus data, without any augmentations, as presented in training data section. In the second experiment (i.e. baseline with augmentations), we added scale and rotation augmentations to the training data. In the third experiment (i.e. PIXSAT) we trained our ship detection method only on PIXSAT ship detection training data and in the last experiment (i.e. baseline + PIXSAT), we initialized the method with model trained on Kaggle Airbus and fine-tuned it on PIXSAT dataset.

Overall results across all ship lengths are reported in table~\ref{results_summary}. With our baseline, trained only on Kaggle Airbus data, we successfully detected approximately 50\% of the ships. Overall the results on Planet data are slightly better due to smaller difference in spatial resolution between training and testing data (1.5m vs. 3m). Interestingly, scale augmentations in our second experiment didn't improve the results as much as expected, according to~\cite{high_low_res}. Results only slightly improved on ESA Sentinel-2 data, which was expected due to much lower spatial resolution. Majority of ship annotations in Kaggle Airbus are on open sea, which is contrary to PIXSAT dataset. Results greatly improve when we train the method on PIXSAT dataset. This is not only due to same spatial resolution and sensor characteristics, but also due to introduction of training data with ships in port area. Fine-tuning our method on PIXSAT data, when initialized on Kaggle Airbus did slighty improve the results on Planet data, which was expected, due to more similar spatial resolution.

\begin{table}[ht!]
\caption{Overall ship detection performance across all ship lengths}
\begin{center}
\begin{tabular}{|c|c|c|c|c|}
\hline
 &\multicolumn{2}{|c|}{\textbf{SF}}&\multicolumn{2}{|c|}{\textbf{LB}} \\
\cline{2-5} 
 & \textbf{Planet}& \textbf{Sentinel}& \textbf{Planet}& \textbf{Sentinel} \\
\hline
\textbf{baseline} & 52\% & 54\% & 54\% & 41\% \\
\hline
\textbf{baseline (aug.)} & 50\% & 56\% & 52\% & 45\% \\
\hline
\textbf{PIXSAT} & 57\% & 86\% & 72\% & 92\% \\
\hline
\textbf{baseline + PIXSAT} & 61\% & 87\% & 76\% & 84\% \\
\hline
\end{tabular}
\label{results_summary}
\end{center}
\end{table}

We also report ship detection rate across different ship lengths, for all mentioned experiments, locations and satellite constellations. Results reported in figures~\ref{baseline_hist}, \ref{baseline_aug_hist}, \ref{pixsat_hist} and \ref{pixsat_finetune_hist} need to be correlated with the number of ships, of particular length, captured in the test part of the PIXSAT dataset and presented in figure~\ref{fig:2016_length_hist}.

We can clearly see that ship detection performance on smaller ships is much worse, compared to bigger ships. In our training data, we focused on ships larger than 50 meters, to avoid noise and due to our interest in commercial ships only, due to availability of AIS data. Majority of ships captured in PIXSAT test set are in the range between 100 and 350 meters. We can see that the introduction of scale augmentations, presented in figure~\ref{baseline_aug_hist}, improved detection rates of smaller ships in the range between 100 and 200 meters. Detection rate did not improve much with larger ships. We can see overall improvement across all ship lengths after the introduction of the PIXSAT dataset in the training procedure. We can also see that the detection rate  slightly increases, with the ship lengths. This was not directly noticeable when trained only on Kaggle Airbus dataset. This is due to presence of such ships in port areas, mostly moored and as such not detected when trained only with Kaggle Airbus data. We can also see that the ship detection performance is in general better with ESA Sentinel-2 satellite, which needs to be considered with caution, as the results cannot be directly compared. The satellite imagery of different satellite constellations is not captured at the same times, as well as we manually filtered test part of the PIXSAT dataset. We filtered out false AIS matches, as well as matches, where ship cannot be clearly recognized by a human annotator. In the case of Planet Dove imagery, with increased spatial resolution of 3m, the resulting dataset is more challenging compared to ESA Sentinel-2. There is also much more ships captured with Planet Dove, as presented in figure~\ref{fig:2016_length_hist}.

\begin{figure*}[htbp]
    \centering
    \begin{subfigure}{1\textwidth}
      \includegraphics[width=\linewidth]{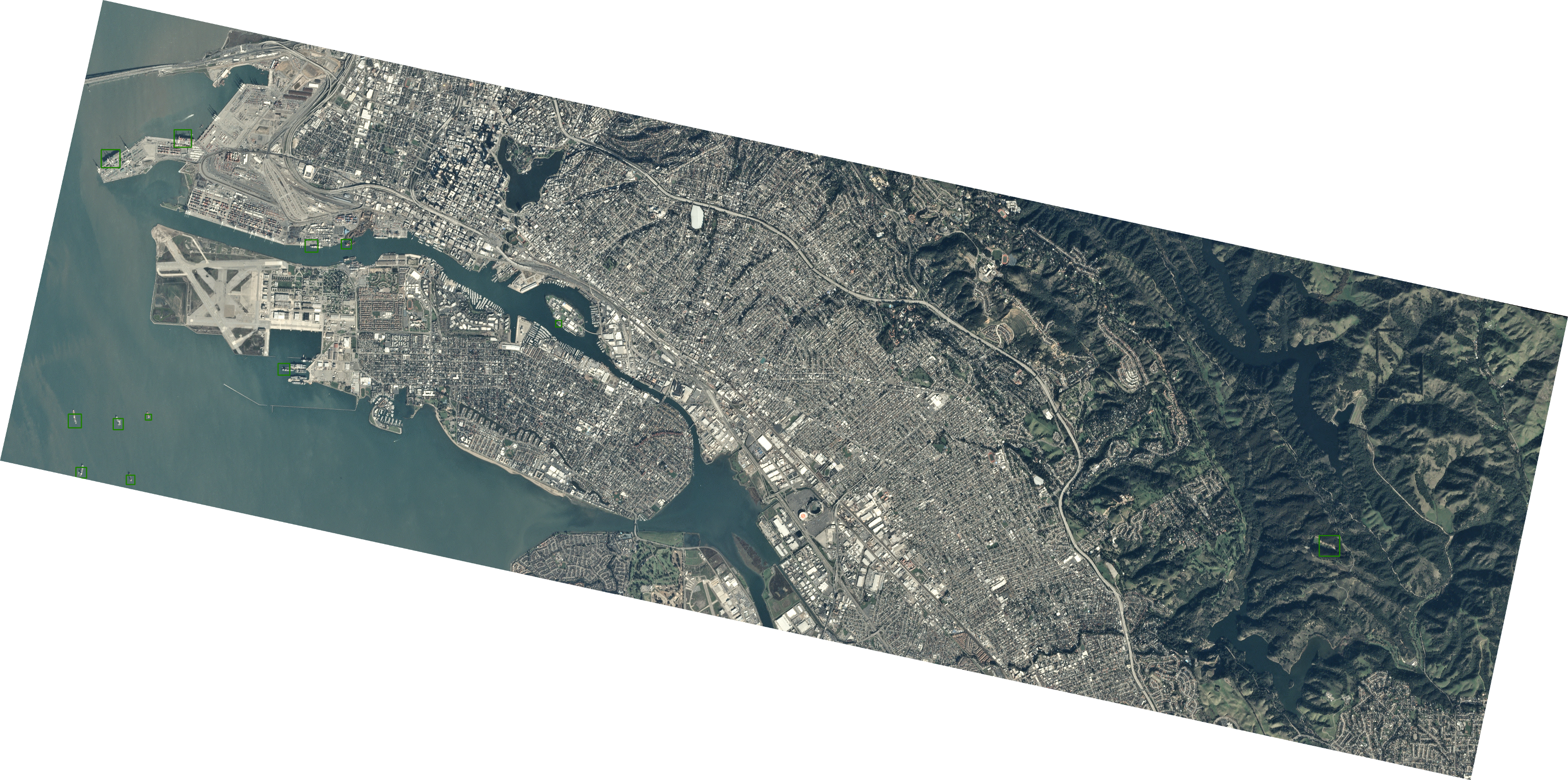}
      \caption{San Francisco Bay area}
      \label{results_planet_sf_image}
    \end{subfigure}\hfil
    \begin{subfigure}{1\textwidth}
      \includegraphics[width=\linewidth]{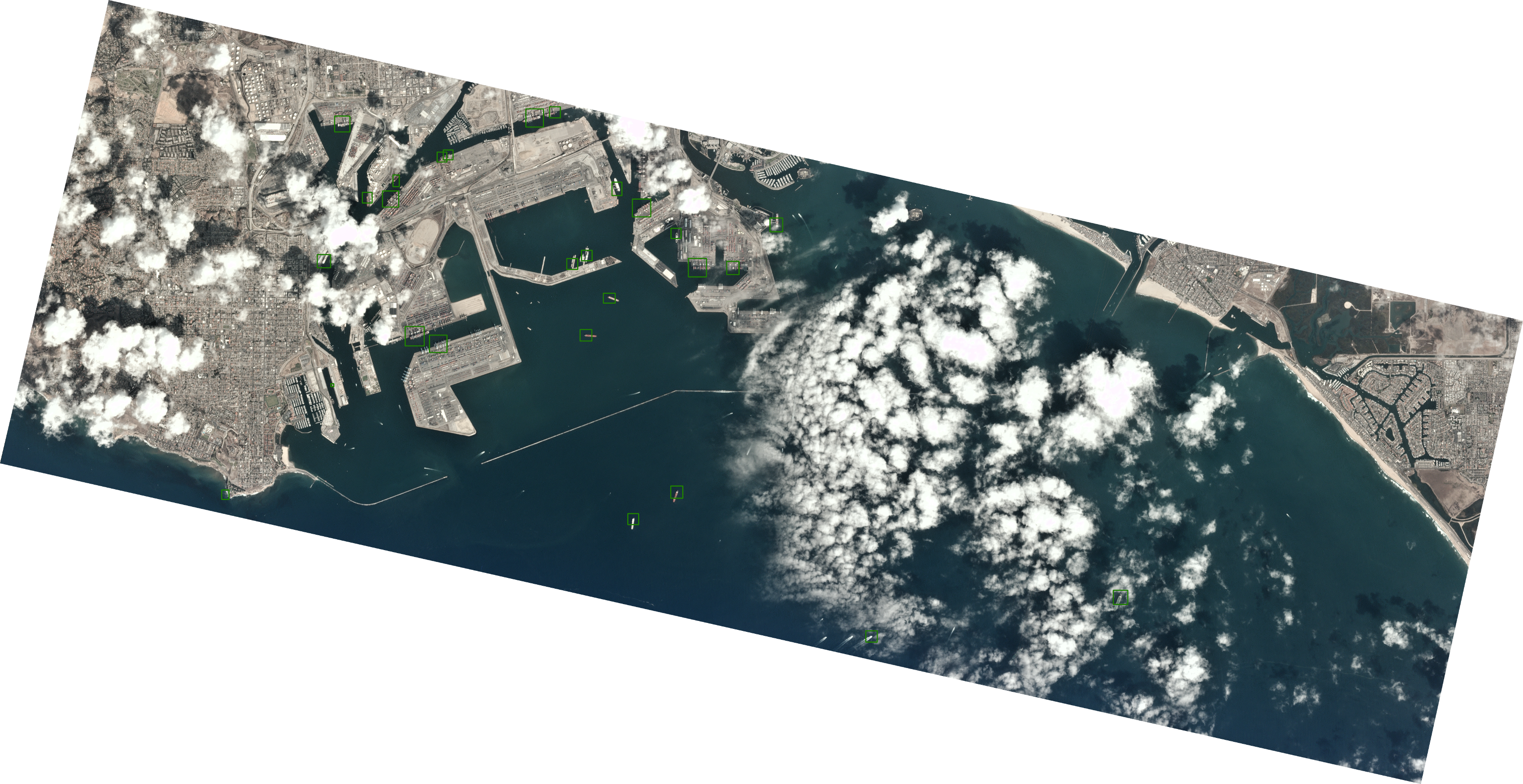}
      \caption{Port of Long Beach area}
      \label{results_planet_lb_image}
    \end{subfigure}\hfil
    \caption{Results of best performing model (baseline + PIXSAT) on Planet Dove satellite imagery\textsuperscript{*}. Best viewed in digital version with zoom.}
    \small\textsuperscript{*} Contains modified Open California Satellite Imagery \textcopyright 2019 Planet Labs Inc. licensed under CC BY-SA 4.0.
\end{figure*}

\begin{figure*}[htbp]
\centering
\centerline{\includegraphics[width=1\textwidth]{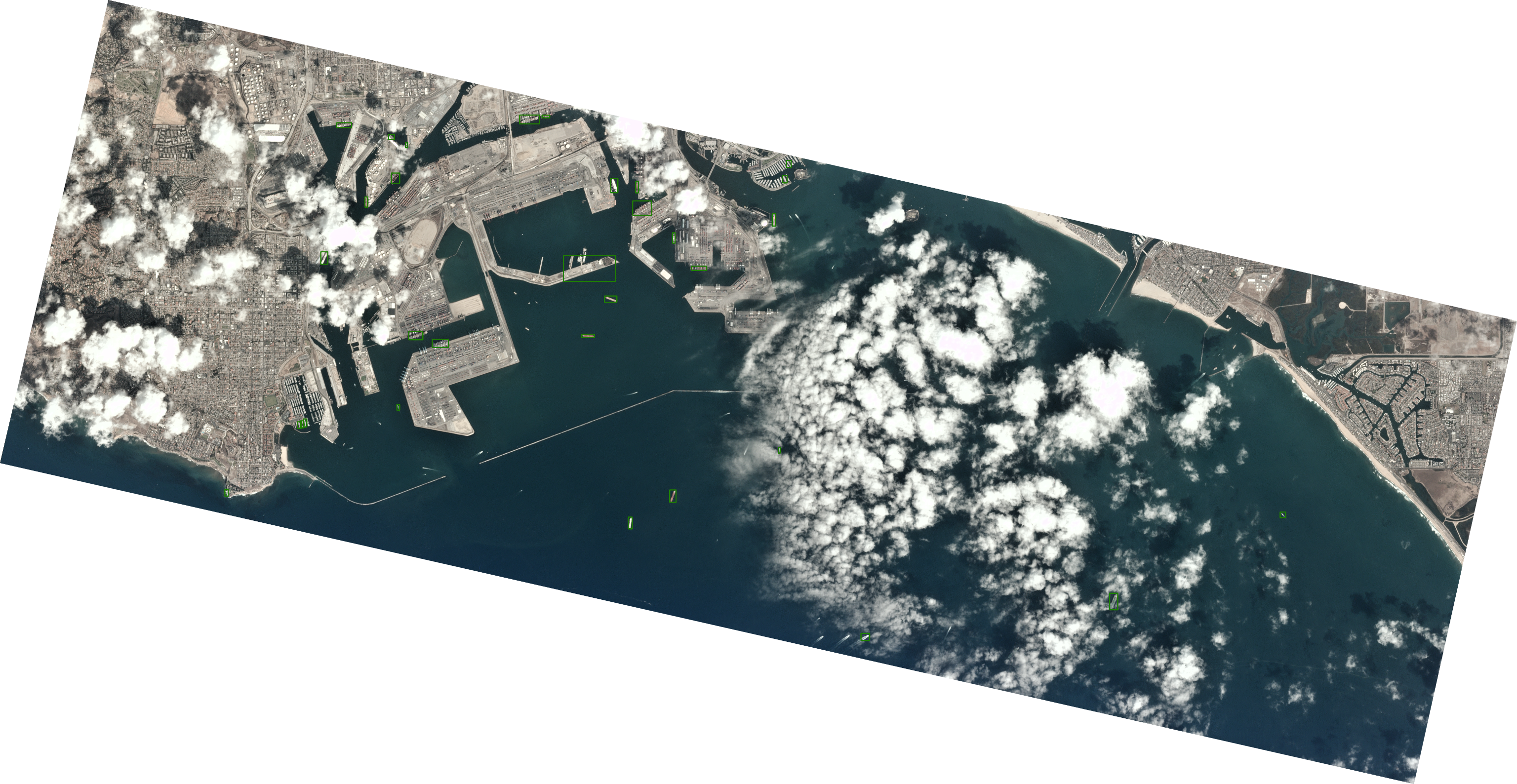}}
\caption{Results of baseline model on Planet Dove satellite imagery of Port of Long Beach\textsuperscript{*}. Best viewed in digital version with zoom.}
\small\textsuperscript{*} Contains modified Open California Satellite Imagery \textcopyright 2019 Planet Labs Inc. licensed under CC BY-SA 4.0.
\label{fig:planet_lb_results_baseline}
\end{figure*}

\begin{figure*}[htbp]
    \centering
    \begin{subfigure}{0.49\textwidth}
      \includegraphics[width=\linewidth]{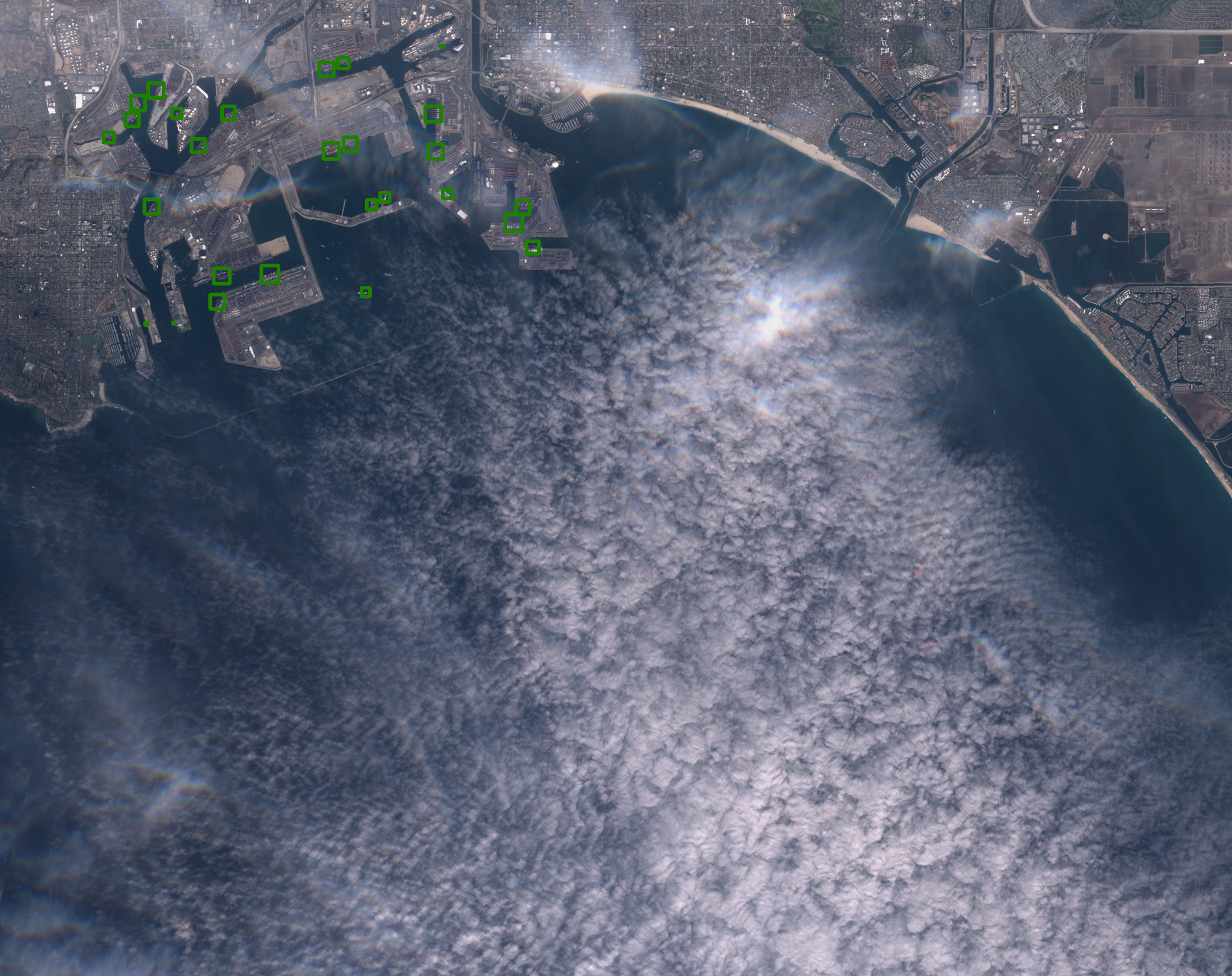}
      \caption{Port of Long Beach area}
      \label{results_sentinel_lb_image}
    \end{subfigure}\hfil
    \begin{subfigure}{0.49\textwidth}
      \includegraphics[width=\linewidth]{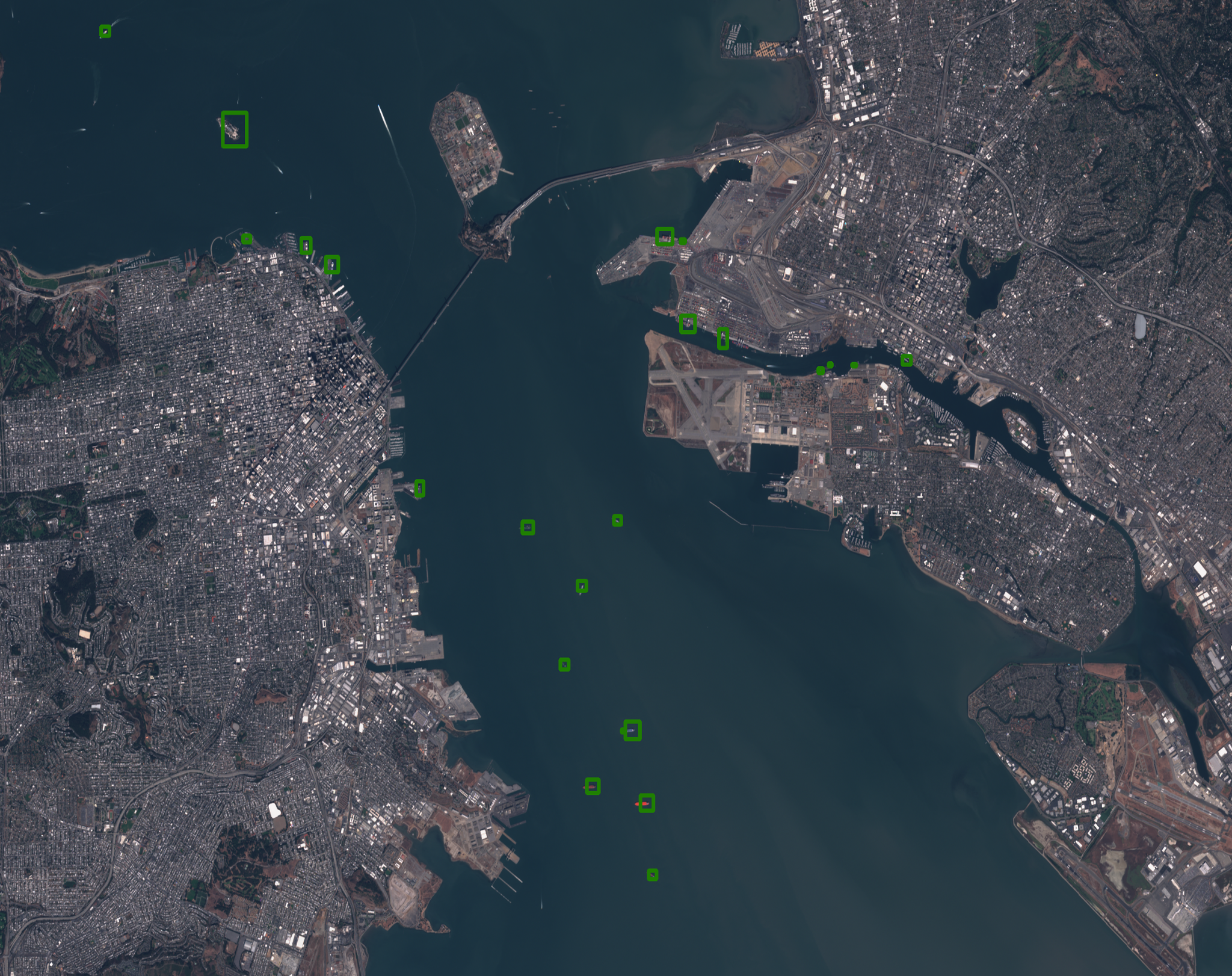}
      \caption{San Francisco Bay area}
      \label{results_sentinel_sf_image}
    \end{subfigure}\hfil
    \caption{Results of best performing model (baseline + PIXSAT) on ESA Sentinel-2 imagery\textsuperscript{*}. Best viewed in digital version with zoom.}
    \small\textsuperscript{*} Contains modified Copernicus Sentinel data from Sentinel Hub licensed under CC BY-NC 4.0.
\end{figure*}

We also present qualitative results on operational satellite imagery, used in the experiments. Figures~\ref{results_planet_lb_image} and~\ref{results_planet_sf_image} present results of the best performing model on Planet Dove satellite imagery. Similar results for ESA Sentinel-2 are presented in figures~\ref{results_sentinel_lb_image} and~\ref{results_sentinel_sf_image}. We can see that detections are robust, in the port area, as well as in the area that is covered with clouds. One can notice, that the detections do not capture the whole area of the ship. This is due to our training data from PIXSAT, where we fitted rectangles of ships lengths to the reported AIS positions, which are usually captured in the ship bridge area. The selection of rectangles size and the influence of background on the detection performance is still to be investigated for future work. Figure~\ref{fig:planet_lb_results_baseline} presents results of the baseline model, which captures ship dimensions much more accurately, but it fails in the port area, due to lack of training imagery in such environment. We have also investigated the use of heading information in AIS data in order to make annotations much more accurate, but we didn't find it reliable enough, especially for stationary ships.

\section{Conclusion}

In this paper we presented a novel use of AIS and satellite imagery data in order to perform ship detection across operational satellite imagery of medium spatial resolution. We presented a large scale ship detection dataset - PIXSAT, which presents a novel approach of using weakly annotated data, which is available in abundance, for developing ship detection methods, without the need of human labeled data. We showed that solely by using weakly annotated data we improve the results compared to using existing, human labeled datasets. We applied the methodology to the largely unused (in maritime domain) medium resolution optical satellite imagery from ESA Sentinel-2 and Planet Dove satellite constellations. The presented methodology is general and applicable also for other satellite constellations and also opens up opportunities for new use cases in the maritime domain.

\section*{Acknowledgment}

This work was partially supported by the European Commission through the Horizon 2020 research and innovation program under grants 769355 (PIXEL), 688201 (M2DC) and 690907 (IDENTITY).

\bibliographystyle{IEEEtran}
\bibliography{IEEEabrv,IEEEexample}

\thanks{
\textcopyright 2019 IEEE. Personal use of this material is permitted. Permission from IEEE must be obtained for all other uses, in any current or future media, including reprinting/republishing this material for advertising or promotional purposes, creating new collective works, for resale or redistribution to servers or lists, or reuse of any copyrighted component of this work in other works.
}

\end{document}